\documentclass{article}

\usepackage{PRIMEarxiv}

\usepackage[utf8]{inputenc} % allow utf-8 input
\usepackage[T1]{fontenc}    % use 8-bit T1 fonts
\usepackage{hyperref}       % hyperlinks
\usepackage{url}            % simple URL typesetting
\usepackage{booktabs}       % professional-quality tables
\usepackage{amsfonts}       % blackboard math symbols
\usepackage{nicefrac}       % compact symbols for 1/2, etc.
\usepackage{microtype}      % microtypography
\usepackage{lipsum}
\usepackage{fancyhdr}       % header
\usepackage{graphicx}       % graphics
\graphicspath{{media/}}     % organize your images and other figures under media/ folder

\usepackage{mathtools}
\usepackage{amsfonts}

\usepackage{graphicx}

\usepackage{multicol}
\usepackage{multirow}

\usepackage{makecell}
\usepackage{arydshln}
\usepackage{enumitem} 
\usepackage{pifont}
\usepackage{comment}
\usepackage{amsmath}
\usepackage{romannum}

\usepackage{braket}

\usepackage{MnSymbol}
\usepackage{adjustbox}

\newtheorem{definition}{Definition}
\newtheorem{proposition}{Proposition}

\newenvironment{proofsketch}{\textbf{Proof Sketch:}}{\hfill {\large$\blacksquare$}}

\title{From Proposal to Execution: Modeling Interaction Semantics in Shared Autonomy 
}

\author{
  Shashank Shekhar, Laurent Jeanpierre, and Abdel-Illah Mouaddib \\
  Université de Caen Normandie, GREYC-CNRS UMR 6072 \\ Caen, France \\ 
  \texttt{\{shashank.shekhar, laurent.jeanpierre, abdel-illah.mouaddib\}@unicaen.fr} \\
}

\begin{document}
\maketitle

\begin{abstract}
% \lipsum[1]
In many human-AI systems, candidate actions are not executed directly but are first evaluated by another decision-making entity under a potentially different objective or decision model.
Standard decision-theoretic frameworks do not require this interaction to be represented explicitly, as they define transition dynamics directly over state-action pairs, allowing evaluation to be absorbed within the transition model.
We introduce a semantics-aware decision framework in which a candidate action is proposed, evaluated under an interaction context, and then resolved into an executed outcome.
We formalize this through a joint action language and an interpretation function that maps each joint action and context to a semantic outcome.
Transition and reward dynamics are induced through these semantic outcomes, thereby introducing an explicit representational layer for \textit{interaction semantics}.
Semantic outcomes provide explicit units for organizing interaction principles, which can be instantiated through application-specific evaluation mechanisms and reused across distinct interaction instances.
This provides a structured interface for specifying context-dependent evaluation separately from the resulting execution dynamics, without requiring changes to the underlying planning algorithm.
We instantiate the framework in shared human-AI autonomy, capturing interaction patterns such as advisory and authorization, and illustrating how interaction context influences decision making.
The resulting formulation remains compatible with standard Markov decision process (MDP) planning.
\end{abstract}

% keywords can be removed
% \keywords{Evaluation before execution \and Decision making under uncertainty \and Shared and adjustable autotomy}

% \section{Introduction}
\section{Introduction}
\label{sec:introduction}

Standard decision-theoretic models typically specify the consequences of selected actions directly through transition dynamics~\cite{sutton1998reinforcement}.
In many settings, however, candidate actions are first evaluated by another decision-making entity before execution. 
The realized outcome therefore depends not only on the proposed action, but also on the interaction through which it is accepted, modified, rejected, or otherwise resolved.

This interaction pattern arises in shared autonomy, human-in-the-loop learning, and oversight settings, where humans or autonomous policies evaluate proposed actions before they affect the environment~\cite{bonner2000adaptive,Hadfield-Menell16,amershi2014power,OrseauA16}.
In general, many interactive decision-making problems distinguish between proposing an action and determining whether, and in what form, it is ultimately realized.

We adopt the following decision-theoretic principle. 
We model interactive decisions through three stages: (1) a candidate action is proposed, (2) evaluated under an interaction context, and (3) resolved into an executed outcome.

The distinction is not one of expressiveness but of \textit{representation}. 
A sufficiently expressive transition model can encode both evaluation and execution within a single dynamics. 
However, doing so conflates two fundamentally distinct stages of decision making: evaluating and resolving a proposed action, and describing how the environment evolves following its execution. 
Evaluation may itself involve distinct decision-making principles, governed by another entity's objectives, knowledge, authority, or interaction context.
When such processes are embedded within transition dynamics, their identification, specification, and modification may become entangled with task-level execution behavior, particularly when common evaluation principles apply across different actions.
Making this separation explicit provides a dedicated interface for representing and managing these interaction processes separately from the resulting execution dynamics, while preserving the underlying planning model.

Existing shared-autonomy and multi-agent formulations capture such interactions through augmented state variables, state-dependent transitions, or supervisory mechanisms (e.g.,~\cite{VanheeJM21}).
While these approaches can explicitly represent supervisory interventions, the evaluation principles shared across different actions and interaction contexts need not be exposed within the decision-process specification.

We address this by introducing a semantic layer that explicitly models the interaction governing how proposed actions become executed actions.
We define a joint action language for interaction-level proposals, building on joint action models~\cite{Littman94}, together with an interpretation function that maps each joint action under an interaction context to a semantic outcome.
Transition and reward dynamics are induced through these semantic outcomes, which determine how proposed actions are resolved before execution.

Our framework specifies where interaction semantics are represented within the decision process, rather than prescribing how the underlying evaluation mechanisms are obtained. 
For a given application, the interpretation over joint actions and interaction contexts determines the semantic outcomes needed to represent its interaction semantics.
These outcomes constitute the semantic layer and are instantiated through domain-specific evaluation mechanisms, while planning continues over the induced dynamics.
Because multiple interaction instances may realize the same semantic outcome, they naturally share a common semantic representation, enabling structural reuse independent of the particular evaluation mechanism.

Our formulation preserves compatibility with standard MDP planning while making interaction semantics explicit.

We instantiate the framework in shared autonomy~\cite{bonner2000adaptive,CostenGHL24}, where interaction is mediated by autonomy and attention, providing a concrete instantiation of interaction semantics through context-dependent evaluation.  
For example, an AI agent may propose an action that is executed only if the human evaluates it as acceptable. 
This structure captures interaction patterns such as advisory and authorization within a unified representation.

\paragraph{}
% representation -> Theory -> Instantiation -> Analysis
In summary, our \textbf{key contributions} are as follows:
\begin{enumerate}
    
    \item We introduce a semantics-aware decision framework with a joint action language and interpretation function
    that map interaction-level actions to semantic outcomes,
    making interaction semantics explicit.

    \item We establish that semantic outcomes induce semantic regimes that enable structured transition and reward dynamics compatible with standard MDP planning.

    \item We instantiate the framework in shared human-AI autonomy, showing how interaction patterns such as advisory and authorization arise as semantic outcomes within a unified interaction representation.

    \item We analyze the structural implications of explicitly representing interaction semantics relative to implicit encoding, and illustrate the framework in a shared-autonomy navigation domain, demonstrating context-dependent evaluation and its influence on
    decision making.    
\end{enumerate}

\section{Abstract Framework}
\label{sec:abstract_framework}
The framework is defined at an abstract level, independently of any particular application-specific evaluation mechanism.
Throughout the paper, we instantiate the interaction context using autonomy and attention to ground the framework in shared autonomy.
The present formulation assumes finite, observable interaction contexts, while extensions to richer settings remain future work.

\subsection{Interaction Context}

We represent the interaction context through a structured set of variables that determine how candidate actions are interpreted within the semantic layer.
This provides an explicit representation of context-dependent interpretation within the setting we consider here.

\begin{definition}[Interaction Context]
Let $L$ denote a finite set of autonomy levels and $At$ a finite set of attention states.  
An \emph{interaction context} is a pair
\[
(lau, att) \in L \times At,
\]
where $lau$ specifies the distribution of control and decision authority between agents, and $att$ is the attention level available for evaluating and responding to a candidate action.
\end{definition}

\subsection{Joint Action Language}

\begin{definition}[Joint Action Language]
Let $A_i$ and $A_j$ denote the sets of interaction-level actions available to two agents.  
The \emph{joint action language} is the set
\[
A_i^j = A_i \times A_j,
\]
whose elements represent primitive agent interactions.
\end{definition}

Joint actions constitute the primitive interaction objects of the framework. They represent interaction-level proposals between agents (e.g., proposing or responding to), but carry no semantics by themselves. 
Their interpretation depends on the interaction context and is determined by the interpretation function introduced next.

\subsection{Joint Action Semantics}
Not all joint actions are meaningful in every context. 
To capture this, we define an interpretation function that assigns a semantic outcome to each joint action under a given context.

\begin{definition}
[Interpretation Function]
\label{def:interpretation}
Let $\Sigma$ be a set of semantic tags ($\sigma$), each representing a distinct semantic outcome for resolving a proposed action into an executed action.
The interpretation function
\[
\Im_{\mathcal{J}} : (A_i \times A_j) \times (L \times At) \rightarrow \Sigma \cup \{\bot\}
\]
maps a joint action and interaction context to a semantic tag, i.e., 
$\Im_{\mathcal{J}}((a_i,a_j),(lau,att)) = \sigma$, 
and returns $\bot$ if the joint action is not valid under the given context.
\end{definition}

Thus, a semantic outcome represents the interaction semantics governing how a proposed action is resolved before execution. 
The evaluation mechanism corresponding to it determines the executed action, which in turn induces the transition, reward, and discounting.
For example, in an advisory setting, a joint action representing an AI suggestion (e.g., \textit{cross-bridge}) may map to a semantic outcome within which the human evaluates the suggestion, for example, from their understanding and preference, and either accepts and executes it or rejects it. 
Under a different context, the same joint action may induce a different outcome, for instance, if attention is low, the suggestion may be ignored, while triggering a fallback behavior.

% This mapping is defined at the level of individual joint action-context pairs, which reflects that semantic outcome is inherently local and context-dependent.

The interpretation function is defined locally over individual joint action-context pairs. 
The structural reuse induced when multiple pairs map to the same outcome (next section).

We say that a joint action-context pair $((a_i,a_j),(lau,att))$ is \emph{well-defined} if
$\Im_{\mathcal{J}}((a_i,a_j),(lau,att)) \in \Sigma,$
and \emph{invalid} otherwise.

\subsection{Semantics-Aware Decision Making}
\begin{definition}[Semantics-Aware Decision Process]
\label{def:main_model}
A semantics-aware decision process is a tuple
\[
\mathcal{P}_i^j = \langle S^{ext}, A_i^j, \Sigma, \Im_{\mathcal{J}}, \{Tr_i^{j,\sigma}, R_i^{j,\sigma}, \gamma_i^{j,\sigma}\}_{\sigma \in \Sigma} \rangle,
\]
where:
\begin{itemize}
    \item $S^{ext} = S \times L \times At$ is the extended state space, where $S$ is the task state space and $s = (x,lau,att) \in S^{ext}$;
    \item $A_i^j = A_i \times A_j$ is the joint action space;
    \item $\Sigma$ is a finite set of semantic outcomes;
    \item $\Im_{\mathcal{J}} : (A_i \times A_j) \times (L \times At) \rightarrow \Sigma \cup \{\bot\}$ is the interpretation function;
    \item for each $\sigma \in \Sigma$, the functions 
    $Tr_i^{j,\sigma}(s' \mid s,(a_i,a_j))$, 
    $R_i^{j,\sigma}(s,(a_i,a_j),s')$, and 
    $\gamma_i^{j,\sigma}(s,(a_i,a_j), s')$ 
    specify the $\sigma$-indexed dynamics.
\end{itemize}

For any $s=(x,lau,att)$ and joint action $(a_i,a_j)$ such that 
$\sigma = \Im_{\mathcal{J}}((a_i,a_j),(lau,att)) \in \Sigma$, 
the induced dynamics are given by
\[Tr_i^j(s' \mid s,(a_i,a_j)) = Tr_i^{j,\sigma}(s' \mid s,(a_i,a_j)),\] 
\[
R_i^j(s,(a_i,a_j),s') = R_i^{j,\sigma}(s,(a_i,a_j),s'), ~\text{and}\]
\[\gamma_i^j(s,(a_i,a_j),s') = \gamma_i^{j,\sigma}(s,(a_i,a_j),s').\]

Thus, $\Im_{\mathcal J}$ maps each joint action and interaction context to the semantic outcome governing how the proposal is resolved, while transition, reward, and discount remain functions of the state and joint action through the corresponding $\sigma$-indexed dynamics.
\end{definition}

\section{Formal Properties}
\label{sec:formal_properties}
% v1

This section characterizes the structure induced by
interaction semantics, focusing on how semantic
outcomes organize the induced transition and reward
dynamics. The discount factor is omitted for clarity.

By Definition~\ref{def:interpretation}, the
interpretation function depends only on the joint
action and interaction context, independently of
the task state $x$. Thus, for a fixed joint action
and context, the same semantic outcome $\sigma$
is selected across task states. However, the
corresponding $\sigma$-indexed transition and reward
functions may still depend on the full extended
state and joint action. Hence, the selection of
an evaluation principle is independent of the
task state, while its induced dynamics remain
state-dependent.

Different joint action-context pairs may induce the same semantic outcome when they are \textit{``indistinguishable''} at the level of interaction semantics. For example, low \textit{att} may map different interactions to recovery mode.

Although $\Im_{\mathcal{J}}$ is defined for individual joint action-context pairs, distinct pairs may share the same semantic outcome. Consequently, the local interpretation induces a global organization of the interaction space.
We refer to the resulting equivalence classes as semantic regimes.

\begin{proposition}[Semantic Regimes]
\label{theory:Theorem_1}
Let 
$\Im_{\mathcal{J}} : (A_i \times A_j) \times (L \times At) \rightarrow \Sigma \cup \{\bot\}$
be an interpretation function.
Define the set of well-defined joint action-context pairs
\[
\mathcal{D} = \{((a_i,a_j),(lau,att)) \mid \Im_{\mathcal{J}}((a_i,a_j),(lau,att)) \in \Sigma\},
\]
and an equivalence relation $\sim$ on $\mathcal{D}$ by
\[
((a_i,a_j),(lau,att)) \sim ((a'_i,a'_j),(lau',att'))
\iff
\Im_{\mathcal{J}}((a_i,a_j),(lau,att))
=
\Im_{\mathcal{J}}((a'_i,a'_j),(lau',att')).
\]

Then:
\begin{enumerate}
    \item The relation $\sim$ partitions $\mathcal{D}$ into equivalence classes indexed by outcomes $\sigma$ in the image of $\Im_{\mathcal{J}}$, given by
    \[
    \mathcal{R}_\sigma 
    =
    \{((a_i,a_j),(lau,att)) \in \mathcal{D} \mid 
    \Im_{\mathcal{J}}((a_i,a_j),(lau,att)) = \sigma\}.
    \]

    \item For any $\sigma \in \Sigma$, all pairs in $\mathcal{R}_\sigma$ are associated with the same semantic outcome $\sigma$. Consequently, they share the same $\sigma$-indexed evaluation mechanism and therefore invoke the same $\sigma$-indexed functions to compute induced dynamics. 
    In particular, for any $((a_i,a_j),(lau,att)) \in \mathcal{R}_\sigma$, the induced transition and reward are    
    \[
    Tr_i^j(\cdot \mid s,(a_i,a_j)) = Tr_i^{j,\sigma}(\cdot \mid s,(a_i,a_j)) \; \textit{and} \] 
    \[ 
    R_i^j(s,(a_i,a_j),s') = R_i^{j,\sigma}(s,(a_i,a_j),s').
    \]    
    
    Thus, the interpretation function organizes distinct
    joint action-context pairs through common
    $\sigma$-indexed evaluation mechanisms. This
    provides an explicit structure for specifying
    interaction semantics at the level of semantic
    outcomes, while allowing the induced transition
    and reward dynamics to vary with the underlying
    state and joint action.
    
    % As a result, transition and reward can be specified at the level of $\sigma$ and reused across all pairs in $\mathcal{R}_\sigma$, without requiring case-by-case specification over joint actions or contexts.
\end{enumerate}

We refer to each class $\mathcal{R}_\sigma$ as a \emph{semantic regime}.
\end{proposition}

\begin{proofsketch}
% The relation $\sim$ is defined by equality under $\Im_{\mathcal{J}}$ and is therefore an equivalence relation on $\mathcal{D}$. 
% Its equivalence classes are the preimages of elements of $\Sigma$ under $\Im_{\mathcal{J}}$, which yields the partition.
%
The relation $\sim$ is defined by equality under $\Im_{\mathcal{J}}$ and is therefore an equivalence relation on $\mathcal{D}$. The resulting equivalence classes are precisely the preimages of elements of $\Sigma$ under $\Im_{\mathcal{J}}$, which form a partition of $\mathcal{D}$.

For the second claim, let two pairs belong to $\mathcal{R}_\sigma$. By definition, both are mapped to the same semantic outcome $\sigma$. 
By Def.~\ref{def:main_model}, the induced transition and reward for each pair are given by the same $\sigma$-indexed functions, establishing that all pairs in $\mathcal{R}_\sigma$ share the same semantic outcome and therefore the same evaluation mechanism for resolving proposed actions before execution. 
The final statement follows directly from this shared structure.
\end{proofsketch}

Thus, we show that interaction semantics organize joint action-context pairs into semantic regimes $\mathcal{R}_\sigma$.
Each regime is associated with a common semantic outcome, instantiated by a common evaluation mechanism for resolving proposed actions, while remaining parameterized by the underlying state and joint action.
% As a result, transition and reward are specified once per semantic outcome $\sigma$ and reused across all pairs assigned to that regime, exposing reusable interaction structure that would otherwise remain implicit in transition definitions.
As a result, transition and reward are specified once per semantic outcome $\sigma$ and reused across all pairs assigned to that regime.
This makes the reusable interaction structure explicit rather than leaving it implicit in transition definitions.

%%%%%%

\section{Semantic Outcome-Based Planning}
\label{sec:planning}

Planning consists of computing an optimal policy over extended states 
$s = (x, lau, att)$.

At any state $s$, admissible joint actions $(a_i,a_j) \in A_i \times A_j$ are those for which
$\Im_{\mathcal{J}}((a_i,a_j),(lau,att)) = \sigma \in \Sigma$.

% We formulate planning over interaction-level joint actions. 
Planning is performed over interaction-level joint actions, which represent candidate interaction proposals rather than directly executed actions.
The maximization over $a=(a_i,a_j)$ selects among admissible interaction proposals rather than directly choosing an executed action.
Each proposal is interpreted under the current interaction context, inducing a semantic outcome $\sigma$ that governs how the proposed action is resolved before execution.
The resulting executed action and task dynamics are therefore determined by the evaluation mechanism associated with $\sigma$, which may incorporate another agent's independent response.

A semantic outcome may be instantiated by an evaluation mechanism that resolves a proposal into multiple possible executions (e.g., acceptance or rejection), each inducing different task dynamics. 
% The corresponding $\sigma$-indexed functions represent the resulting expected transition, reward, and discount.

When an evaluation mechanism admits multiple possible
resolutions, the resulting transitions, rewards, and
discounting may depend on the realized resolution.
The induced dynamics marginalize over these internal
resolutions. In particular, the $\sigma$-indexed
discount $\gamma_i^{j,\sigma}(s,a,s')$ represents
the effective discount conditional on the successor
state $s'$, accounting for possible dependence
between the evaluation resolution and the
subsequent transition.

The Bellman optimality equation is:
\begin{equation}
\label{eq:vi_sigma}
V(s) =
\max_{\substack{
a \in A_i^j \ \text{s.t.} \
\sigma = \Im_{\mathcal J}(a,(lau,att)) \neq \bot
}}
\sum_{s'}
Tr_i^{j,\sigma}(s'\mid s,a)
\left[
R_i^{j,\sigma}(s,a,s')
+
\gamma_i^{j,\sigma}(s,a,s')V(s')
\right].
\end{equation}

\begin{proposition}
The semantics-aware decision process $\mathcal{P}_i^j$ induces a Markov decision process over $S^{ext}$ when restricted to well-defined joint action-context pairs.
\end{proposition}

\begin{proofsketch}
Let $s=(x,lau,att)\in S^{ext}$ and $(a_i,a_j)\in A_i \times A_j$ be such that 
$\Im_{\mathcal{J}}((a_i,a_j),(lau,att)) = \sigma \in \Sigma$.

By definition, the induced transition, reward, and discount  are given by
$Tr_i^j(\cdot \mid s,(a_i,a_j)) = Tr_i^{j,\sigma}(\cdot \mid s,(a_i,a_j)), \quad
R_i^j(s,(a_i,a_j),\cdot) = R_i^{j,\sigma}(s,(a_i,a_j),\cdot),$
and $\gamma_i^j(s,(a_i,a_j),.) = \gamma_i^{j,\sigma}(s,(a_i,a_j),.)$.

As $\sigma$ is uniquely determined by $((a_i,a_j), (lau,att))$, and the evaluation mechanism associated with $\sigma$ induces transition, reward, and discount functions that depend only on the current state $s$ and joint action $(a_i,a_j)$, the induced dynamics depend only on $(s,(a_i,a_j))$ and not on prior history.

Therefore, the process satisfies the Markov property and defines a standard MDP over $S^{ext}$.
\end{proofsketch}

Planning is performed over admissible joint actions, assumed non-empty for all states (else the state is treated as terminal). 
The semantic outcome indexes the evaluation mechanism through which transition, reward, and discount are induced, while the planning problem itself remains defined over the original extended state and joint action spaces.

%%%%%%%%%%%%%%
% \section{A Grounded Semantics-Aware Decision Framework}
\section{Shared Autonomy Instantiation}
\label{sec:grounded_framework}

\subsection{Baseline: Direct Execution}
Consider two decision processes over the (shared) task state space $x \in S$:
\[
\textit{(AI)} \quad AI_i = \langle S, A_i^{task}, Tr_i, R_i, \gamma_i \rangle, 
\quad
\textit{(Human)} \quad H_j = \langle S, A_j^{task}, Tr_j, R_j, \gamma_j \rangle.
\]

Here, $A^{task}$ denotes task-level actions, while $Tr$, $R$, and $\gamma$ capture agent-specific transition, reward, and discounting models. 
Differences in these components reflect variations in capability, preferences, and decision-making.

Standard modes arise as special cases: \textit{teleoperation}, where the human selects and executes actions using $(Tr_j,R_j)$; \textit{full autonomy}, where the AI executes actions using $(Tr_i,R_i)$; and \textit{mixed control}, where execution alternates between agents. In all cases, selected actions are executed directly, without an explicit intermediate evaluation process~\cite{MouaddibZBJ10}.

\begin{table}[t]
\centering
\begin{adjustbox}{max width=0.92\columnwidth}
\begin{tabular}{p{0.39\linewidth}|p{0.55\linewidth}}
\textbf{Action Type} & \textbf{Description} \\ \hline

\textit{Execute (Full Autonomy)} 
& Execute a candidate task action $a_c \in A^{task}$ directly. \\ \hdashline

\textit{Advise} 
& Suggest a task action $a_c$ without enforcing execution.\\ \hdashline

\textit{Request Authorization} 
& Propose $a_c$ and request approval before execution. \\ \hdashline

\textit{Noop} 
& Remain passive in the interaction. \\
\hline
\end{tabular}
\end{adjustbox}
% \caption{Interaction-level actions available to the AI/robot. Actions may refer to a candidate task action $a_c \in A^{task}$.}
\caption{Interaction-level actions for the AI/robot, referencing a candidate task action $a_c \in A^{task}$.}
\label{tab:robot_actions_new}
\end{table}

\begin{table}[t]
\centering
\begin{adjustbox}{max width=0.92\columnwidth}
\begin{tabular}{p{0.39\linewidth}|p{0.55\linewidth}}
\textbf{Action Type} & \textbf{Description} \\ \hline

\textit{Execute (Teleoperation)} 
& Directly execute a candidate action $a_c \in A^{task}$ via teleoperation. \\ \hdashline

\textit{Authorize}
& Engage in evaluation of a proposed candidate action $a_c$. \\ \hdashline

\textit{Request Advice} 
& Actively query AI for guidance. \\ \hdashline

\textit{Noop} 
& Remain passive in the interaction. \\ \hline
\end{tabular}
\end{adjustbox}
\caption{Interaction-level actions available to the human.}
\label{tab:human_actions_new}
\end{table}

\subsection{Interaction-Level Actions and Joint Actions}

We introduce interaction-level actions through which agents propose, respond to, or execute candidate task actions, $a_c \in A^{task}$~ (Tables~\ref{tab:robot_actions_new} \& \ref{tab:human_actions_new}). 
An interaction-level action may refer to $a_c$ but does not directly induce state transitions.

% ----

A joint action captures a primitive interaction-level proposal. 
We restrict focus to meaningful joint actions, defined as semantically coherent pairs referring to a single $a_c$.

Canonical patterns are \textit{human execution:} $(\text{Noop}, \text{Execute}(a_c))$, 
\textit{robot execution:} $(\text{Execute}(a_c), \text{Noop})$, and 
\textit{interaction:} $(\text{Propose}(a_c), \text{Respond}(a_c))$, 
where \textit{Propose} and \textit{Respond} denote abstract interaction roles instantiated by concrete actions (e.g., \textit{Advise}, \textit{Request Authorization} from the tables). 
These patterns subsume both direct execution and interactive decision making. 
Joint actions specify proposals and interaction roles, while execution is resolved through semantic outcomes under the interaction context.

\subsection{Semantic Outcomes and Induced Dynamics}
We now provide one application-specific instantiation of semantic outcomes through evaluation mechanisms.

For a state $s=(x,lau,att)$ and a joint action $(a_i,a_j)$ associated with a candidate action $a_c$, each semantic outcome is instantiated by an evaluation mechanism that resolves the proposal into an executed action $a^{\mathrm{exec}}$. 
The resulting execution induces the corresponding transition, reward, and discount dynamics.

For exposition, we focus on transition dynamics, while reward and discounting follow analogously.

The transition jointly updates the task state and interaction context,
$
Tr_i^{j,\sigma}(s' \mid s,(a_i,a_j)), \ \text{where} \ s'=(x',c').
$

When the mapping from $a_c$ to $a^{\mathrm{exec}}$ is stochastic, the induced transition marginalizes over the possible executed actions:
\[
Tr_i^{j,\sigma}(s' \mid s,(a_i,a_j))
=
\mathbb{E}_{a^{\mathrm{exec}}}
\!\left[
Tr_i^{j,x}(x' \mid s,a^{\mathrm{exec}},\sigma)\,
Tr_i^{j,c}(c' \mid s,a^{\mathrm{exec}},\sigma)
\right],
\]
where $s'=(x',c')$, $c'=(lau',att')$, and the expectation is taken over the executed action induced by the evaluation mechanism associated with $\sigma$. This defines the $\sigma$-conditioned transition kernel used in planning (Eq.~\ref{eq:vi_sigma}).

\vspace{0.3em}
\noindent 
{\normalsize
\textbf{Representative semantic outcomes.}}

% The semantic outcomes below are 
They are illustrative rather than exhaustive, representing one possible instantiation of interaction semantics within the semantic layer introduced in the abstract framework.

For clarity, we show only task-state transitions, while context dynamics are defined analogously.

\noindent
\textbf{(1) Teleoperation ($\sigma_{\mathrm{TO}}$).}  
The human executes the action $a_c$:
\[
a^{\mathrm{exec}} = a_c, \quad
Tr_i^{j,x}(\cdot \mid s,a_c,\sigma_{\mathrm{TO}})
= \widetilde{Tr}_j(\cdot \mid x,a_c,att) \approx Tr_j(\cdot \mid x,a_c).
\]

This recovers standard teleoperation, with $att$ optionally modulating execution fidelity.
% (e.g., degraded control under low attention).

\noindent
\textbf{(2) Full autonomy ($\sigma_{\mathrm{FA}}$).}  
The robot executes action $a_c$:
\[
a^{\mathrm{exec}} = a_c, \quad
Tr_i^{j,x}(\cdot \mid s,a_c,\sigma_{\mathrm{FA}})
= \widetilde{Tr}_i(\cdot \mid x,a_c,att) \approx Tr_i(\cdot \mid x,a_c).
\]

Here, attention may capture supervision or monitoring effects influencing execution reliability.

\noindent
\textbf{(3) Advisory interaction ($\sigma_{\mathrm{adv}}$).}  
The robot proposes $a_c$, which is evaluated by the human and accepted with a probability $\rho(s,a_c)$ (see \ref{sec:human_model}). 
The fallback action is the human policy $\pi_j$:
% \sh{verify this factor in the code!}
\[
a^{\mathrm{exec}} \sim 
\begin{cases}
a_c, & \text{with probability } \rho(s,a_c),\\
a' \sim \pi_j(\cdot \mid s), & \text{otherwise}.
\end{cases}
\]

The resulting task transition is,
$
Tr_i^{j,x}(\cdot \mid s,a^{\mathrm{exec}},\sigma_{\mathrm{adv}})
=
\widetilde{Tr}_j(\cdot \mid x,a^{\mathrm{exec}},att).
$

Thus, evaluation determines whether the proposed action is executed or replaced by another action, while the transition depends on the current state and the executed action.

The same semantic outcome is associated with reward and discount through the same evaluation mechanism.
In our instantiation, the effective discount factor may depend on the evaluation outcome: accepted proposals are associated with a longer effective planning horizon, whereas fallback behavior follows the human's default horizon.
% This illustrates one possible evaluation model rather than a property of the framework itself.

\noindent

\textbf{(4) Authorization interaction ($\sigma_{\mathrm{auth}}$).}  
Execution follows the same acceptance rule as $\sigma_{\mathrm{adv}}$. 
In this instantiation, we consider that, if the proposal is rejected, the control transfers to the human, who selects the action to execute according to their own policy $\pi_j$, rather than reconsidering $a_c$. 
Thus, the resulting transition depends on the executing agent:
\[
Tr_i^{j,x}(\cdot \mid s,a^{\mathrm{exec}},\sigma_{\mathrm{auth}})
=
\begin{cases}
\widetilde{Tr}_i(\cdot \mid x,a^{\mathrm{exec}},att), & \text{if accepted},\\
\widetilde{Tr}_j(\cdot \mid x,a^{\mathrm{exec}},att), & \text{otherwise}.
\end{cases}
\]

\noindent
\textbf{(5) Recovery ($\sigma_{\mathrm{rec}}$).}  
Fallback behavior is triggered under failed interaction (e.g., low attention):
\[
a^{\mathrm{exec}} \sim \pi_{\mathrm{rec}}(\cdot \mid s).
\]

The transition follows a recovery policy, reflecting safety mechanisms or external intervention. 
Unlike advisory or authorization instantiations, this is not driven by deliberate evaluation but by breakdown of interaction.

\subsection{Human Evaluation Model}
\label{sec:human_model}
For the advisory and authorization semantic outcomes considered in our experiments, we instantiate the corresponding human evaluation mechanism using a context-modulated one-step lookahead functional, motivated by~\cite{asmar2022collaborative,lu2024does}. This is \textit{one} possible realization of the semantic outcome and is not part of the abstract framework.

{\small
\[
Q_j(s,a_c)
=
\mathbb{E}_{x' \sim \widetilde{Tr}_j(\cdot \mid x,a_c,att)}
\Big[
\widetilde{R}_j(x,a_c,att,x')
+ \gamma_j^{\mathrm{eff}}(att)\, \widehat{V}_j(x')
\Big],
\]
}where $\widehat{V}_j$ is a fixed task-level value approximation, and $\gamma_j^{\mathrm{eff}}(att)$ captures attention-dependent foresight.

The resulting acceptance probability under a stochastic decision rule (e.g., softmax) is
{\small
\[
\rho(s,a_c) = \Pr\!\left(a_c \text{ is selected under the decision rule induced by } Q_j\right).
\]
}

% \section{Interaction Semantics: Evaluation and Decision Behavior}
\section{Qualitative Analysis of Interaction Semantics}
\label{sec:case_study}

\subsection{Explicit Semantics vs. Implicit Encoding}
\noindent\textbf{Standard MDP encoding.}
Interaction behaviors, such as accepting or rejecting
a proposed action following its evaluation under an
interaction context, can be encoded directly within
transition probabilities
$Tr_i^j(s' \mid s,(a_i,a_j))$.
In such encodings, evaluation and execution may be
combined within the induced dynamics rather than
represented as separate processes.

For example, an advisory transition may combine
the probability of accepting a proposal with the
execution dynamics of the accepted or fallback action.
When similar evaluation principles govern different
actions or contexts, their effects may be distributed
across multiple state-action transition definitions.
In the worst case, an unstructured tabular encoding
requires specifying transition and reward dynamics
for up to $|S^{ext}|\cdot|A_i\times A_j|$
state-action cases.

This presents two potential limitations:
(1) \textit{Limited explicit access:} identifying and
isolating common evaluation rules may require
examining multiple transition definitions, and
(2) \textit{Modification cost:} changing evaluation
behavior, e.g., based on attention, may require
revisiting several state-action cases.
Although a suitably factored MDP can avoid these
limitations, identifying such a factorization requires
the modeler to distinguish evaluation from execution
and recognize shared evaluation principles within
potentially intertwined dynamics.

\noindent\textbf{Semantics-aware encoding.}
In contrast, our formulation makes this separation
an explicit modeling principle. The interpretation
function $\Im_{\mathcal J}$ identifies the semantic
outcome governing a joint action under its interaction
context, while the corresponding evaluation mechanism
determines how the proposal is resolved and induces
the resulting dynamics.

This provides a dedicated representational interface
for specifying evaluation principles separately from
task-level execution behavior. When distinct
interactions share a semantic outcome, their
evaluation behavior can be specified through a
common mechanism while retaining state- and
action-dependent execution dynamics.

Consequently, modifying an evaluation mechanism,
e.g., $\rho(s,a_c)$, can update the induced behavior
of the corresponding interactions without independently
redefining their transition specifications.
The underlying execution models can remain unchanged.

The distinction therefore concerns the organization
and modification of interaction models, rather than
the computational complexity of planning itself.
Making evaluation mechanisms explicit allows their
structure to be accessed and managed directly,
instead of requiring its identification and separation
from the induced state-action dynamics.

%%%%%%%%%%%%%%

\subsection{Structural Reuse Through Semantic Outcomes}

Making interaction semantics explicit does not by itself reduce specification effort.
In principle, one could define a distinct mechanism for every joint action-context pair, and we now show that semantic outcomes avoid this redundancy.

Interaction semantics partition joint action-context pairs into regimes $\mathcal{R}_\sigma = \{((a_i,a_j),(lau,att)) \mid \Im_{\mathcal{J}}((a_i,a_j),(lau,att)) = \sigma\}$, each indexed by $\sigma$ (Theorem~\ref{theory:Theorem_1}). 
Without this structure, transition and reward must be specified for each $(s,(a_i,a_j)) \in S^{ext} \times (A_i \times A_j)$. 
With semantic outcomes, the evaluation mechanism is specified once per $\sigma$ and reused across all pairs in $\mathcal R_\sigma$.

% v1
% Interaction semantics partition joint action-context pairs into regimes, $\mathcal{R}_\sigma = \{((a_i,a_j),(lau,att)) \mid \Im_{\mathcal{J}}((a_i,a_j),(lau,att)) = \sigma\}$,
% where each regime corresponds to a semantic outcome $\sigma$ (Theorem~\ref{theory:Theorem_1}).

% The reduction is therefore in the number of distinct mechanisms: each regime $\mathcal{R}_\sigma$ induces a single reusable mechanism.
The reduction is therefore in the number of distinct evaluation mechanisms that must be specified: each regime $\mathcal{R}_\sigma$ shares a single reusable mechanism.

\textit{(Worst Case.)}~ $|\Sigma| = |A_i \times A_j| \cdot |L \times At|$, thus no reduction. 
% However, in structured settings, many pairs share an outcome, so $|\Sigma| \ll |A_i \times A_j| \cdot |L \times At|$, enabling compact specification.
In structured settings, however, many pairs share an outcome, so $|\Sigma| \ll |A_i \times A_j| \cdot |L \times At|$, enabling compact specification by localizing interaction behavior to reusable evaluation mechanisms associated with semantic outcomes rather than distributing it across transition definitions.

\subsection{Bridge Crossing Case Study}

% We illustrate 
We explain the framework by a decision problem with three actions (bridges), each yielding an immediate reward and a probabilistic long-term payoff. 
\textit{B1} (\textit{safe}) gives $15$ (both agents) followed by $10$ with probability $0.8$. 
\textit{B2} (\textit{hidden optimal}) gives $-15$ (human), $-40$ (robot), followed by $150$ with probability $0.99$. 
\textit{B3} (\textit{robot-preferred}) gives $20$ (robot), $-15$ (human), followed by $112$ with probability $0.9$.
The human and robot differ in their foresight: $\gamma_j = 0.2$ and $\gamma_i = 0.99$.

\vspace{0.1cm}
% \paragraph{Baselines.}
\noindent \textbf{Baselines.} \
Under direct execution (no interaction), teleoperation selects B1 with
\[V \approx 15 + 0.2 \cdot 0.8 \cdot 10 = 16.6,\] 
while full autonomy selects B3 with 
\[V \approx 20 + 0.99 \cdot 0.9 \cdot 112 \approx 120.\]

\vspace{0.1cm}
% \paragraph{Advisory interaction (context-dependent evaluation).}
\noindent \textbf{Advisory.} \ Under the advisory evaluation mechanism (\textit{ref. $\sigma_{\mathrm{adv}}$}), the robot proposes to take B2. The human evaluates this proposal with acceptance probability $\rho(s,a_c)$.

If accepted, the human executes and the outcome follows \textit{long-horizon} evaluation,
\[
V_{\text{accept}} \approx -15 + 0.99 \cdot 0.99 \cdot 150 \approx 132.\]
If rejected, the human teleoperates, and it falls back to B1,
\[
V_{\text{reject}} \approx 16.6.
\]
Thus,
\[
V = \rho(s,a_c)\cdot 132 + (1-\rho(s,a_c)) \cdot 16.6.
\]
Note that when a proposal is accepted, the resulting outcome reflects long-horizon reasoning, consistent with the $\sigma$-dependent discounting (as per the evaluation mechanism).

For high attention (suppose $\rho(s,a_c) \approx 0.95$), 
% \[
$V \approx 126$,
% \]
which exceeds both baselines. While, for low attention (suppose $\rho(s,a_c) \approx 0.2$), 
% \[
$V \approx 40$.
% \]

Under the illustrative regime-dependent valuation,
the expected advisory value is
\[
V_{\mathrm{adv}}
= \rho(s,a_c)\cdot 132
+ (1-\rho(s,a_c))\cdot 16.6.
\]
Taking the specified regime-dependent values as the criterion for comparing interaction alternatives, advisory exceeds the full-autonomy reference value of $120$ when
\[
\rho(s,a_c) \geq
\frac{120-16.6}{132-16.6}
\approx 0.90.
\]
Thus, under the adopted valuation scheme, advisory is preferable to full autonomy when the acceptance probability exceeds this threshold.
This illustrates how context-dependent evaluation and heterogeneous decision models can influence the relative desirability of interaction regimes.

\vspace{0.1cm}
\noindent \textit{(Observation.)} \
The optimal decision depends on context (here, attention) through the explicit evaluation mechanism $\rho(s,a_c)$. 
In standard MDPs, this dependence is implicit in transition probabilities, whereas here it is directly modeled. 
% As a result, the policy selects advisory only under sufficient attention, yielding interpretable, context-dependent behavior.
As a result, the value of advisory depends explicitly on the evaluation mechanism instantiating the advisory semantic outcome under the current interaction context, yielding interpretable context-dependent behavior.

%%%%%%%%%%%%%%TABLES
\begin{table}[t]
\centering
\resizebox{0.43\textwidth}{!}{
\begin{tabular}{lccccc}
\hline
$att$ & $\rho(s,a_c)$ & $V^*(s)$ & $best\_act$ & $\Delta Q$ & $Q(s,noop)$ \\
\hline
very low  & 0.504 & 1.876 & right & 1.188 & 0.688 \\
low       & 0.509 & 3.546 & right & 1.355 & 2.192 \\
medium    & 0.519 & 5.059 & right & 1.506 & 3.553 \\
high      & 0.536 & 6.231 & right & 1.623 & 4.608 \\
very high & 0.562 & 7.313 & right & 1.731 & 5.582 \\
\hline
\end{tabular}
}
\caption{Start state $(0,0)$. The reference action ($a_c$ = \textit{right}) remains the optimal advice across all attention levels. Advisory consistently outperforms silence despite only modest growth in acceptance.}
\label{tab:start-lau2}
\end{table}

\begin{table}[t]
\centering
\resizebox{0.43\textwidth}{!}{
\begin{tabular}{lccccc}
\hline
$att$ & $\rho(s,a_c)$ & $V^*(s)$ & $best\_act$ & $\Delta Q$ & $Q(s,noop)$ \\
\hline
very low  & 0.559 & 5.357 & down & 1.536 & 3.821 \\
low       & 0.717 & 8.161 & down & 1.816 & 6.345 \\
medium    & 0.855 & 10.112 & down & 2.011 & 8.101 \\
high      & 0.922 & 11.586 & down & 2.159 & 9.427 \\
very high & 0.916 & 12.773 & down & 2.277 & 10.496 \\
\hline
\end{tabular}
}
\caption{State above bridge \textit{B1} $(2,2)$. The reference action ($a_c$ = \textit{down}) is consistently optimal, while acceptance increases strongly with attention, reflecting increasing agreement between the robot's advice and the human's evaluation.}
\label{tab:above-b1-lau2}
\end{table}

%%%%%%
\begin{table}[t]
\centering
\resizebox{0.43\textwidth}{!}{
\begin{tabular}{lccccc}
\hline
$att$ & $\rho(s,a_c)$ & $V^*(s)$ & $best\_act$ & $\Delta Q$ & $Q(s,noop)$ \\
\hline
very low  & 0.469 & 3.468 & left & 0.979 & 2.121 \\
low       & 0.567 & 5.409 & left & 1.261 & 3.868 \\
medium    & 0.701 & 7.087 & left & 1.684 & 5.378 \\
high      & 0.819 & 10.353 & down & 2.035 & 8.317 \\
very high & 0.856 & 14.043 & down & 2.404 & 11.638 \\
\hline
\end{tabular}
}
\caption{State above bridge \textit{B2} $(4,2)$. The reference action ($a_c$ = \textit{down}). 
A phase transition occurs: the optimal advice changes from \textit{left} to \textit{down} as attention increases, illustrating context-dependent evaluation and execution.}
\label{tab:above-b2-lau2}
\end{table}

\begin{table}[t]
\centering
\resizebox{0.43\textwidth}{!}{
\begin{tabular}{lccccc}
\hline
$att$ & $\rho(s,a_c)$ & $V^*(s)$ & $best\_act$ & $\Delta Q$ & $Q(s,noop)$ \\
\hline
very low  & 0.540 & 11.541 & down & 2.154 & 9.387 \\
low       & 0.587 & 13.980 & down & 2.398 & 11.582 \\
medium    & 0.637 & 16.665 & down & 2.667 & 13.999 \\
high      & 0.677 & 19.573 & down & 2.957 & 16.616 \\
very high & 0.687 & 22.643 & down & 3.264 & 19.379 \\
\hline
\end{tabular}
}
\caption{State in the \textit{B2} corridor $(4,4)$. The reference action ($a_c$ = \textit{down}). The optimal advice remains \textit{down}, while both the advisory advantage over silence and the resulting state value increase with attention.}
\label{tab:b2-corridor-lau2}
\end{table}
%%%%%%%%%%%%%%

% \newtext{
\section{Empirical Evaluation}
\label{sec:experiments}
% 
% \subsubsection{Environment and Design Rationale}
\paragraph{Environment and Design Rationale}
We instantiate the framework in the navigation domain introduced in the case study to examine the interrelation  between autonomy, attention, and human-robot capability asymmetry.

The environment is an $8\times9$ grid with three bridges leading to isolated corridors. Crossing a bridge commits the agent to a corridor, making the initial decision irreversible.

Bridge crossing and subsequent navigation are stochastic and attention-dependent, allowing attention to influence both immediate execution and long-term decision value.

This design isolates the effect of interaction regimes (teleoperation, advisory, and full autonomy) and attention while minimizing confounding navigation effects.

% \subsubsection{Structural Reuse in the Grid World}
% 
\paragraph{Structural Reuse in the Grid World}
The previous section established that semantic outcomes enable reuse of evaluation mechanisms. 
Here we illustrate this property in the implemented domain.

The semantic outcome w.r.t. advisory is instantiated through a single attention-dependent evaluation mechanism, where human acceptance $\rho(s,a_c)$ depends on attention and the estimated quality of the proposed action. 
At $(2,2)$, acceptance increases from very low ($att=1$) to high attention ($att=4$) (\textit{ref.} Table~\ref{tab:above-b1-lau2}).

Because this mechanism is specified once at the semantic outcome level, modifying the acceptance model changes advisory behavior globally without modifying the planner, transition model, or individual state-action definitions.

\subsection{Experimental Results}
Our experiments evaluate one representative instantiation of the semantic layer rather than the general framework itself.

We evaluate our framework at four representative task states corresponding to distinct phases of the navigation task: 
The state $(0,0)$, the states immediately above bridges \textit{B1} $(2,2)$ and \textit{B2} $(4,2)$, and a representative state inside the second corridor $(4,4)$. 
For each state, we fix the interaction context to the advisory regime, while varying the attention level from $1$ to $5$. 
For a fixed \textit{reference action} $a_c$, we report its acceptance probability $\rho(s,a_c)$, the converged optimal state value $V^*(s)$, the optimal advised action selected by planning ($best\_act$), the marginal value of advisory $\Delta Q = Q(s,a_c)-Q(s,\textit{noop})$, and the no-advice baseline $Q(s,\textit{noop})$ (Tables~\ref{tab:start-lau2} to \ref{tab:b2-corridor-lau2}). The main observations are summarized below.

\paragraph{Observation~1: Advisory improves performance even under modest acceptance.}
At the \textit{start state}, the reference action (\textit{right}) remains the optimal advice across all attention levels. 
Although acceptance increases only gradually, advisory consistently outperforms silence, indicating that even modest agreement can improve long-term value when the suggested action aligns with the optimal trajectory.

\paragraph{Observation~2: Acceptance reflects state-dependent agreement.}
Above \textit{B1}, it exhibits the strongest growth in acceptance while consistently selecting \textit{down} as the optimal advice. 
This indicates that acceptance depends jointly on attention and the task state, increasing most rapidly when the proposed action agrees with the human's assessment.

\paragraph{Observation~3: Phase transition at bridge \textit{B2}.}
The region around bridge \textit{B2} exhibits the richest behavior. 
At low attention, the optimal advice is \textit{left}, but as attention increases it switches to \textit{down}. 
(I.e. at low attention, one should not suggest \textit{down} because neither planner nor the human thinks it is a good idea.)
This transition results from the interaction between the evaluation mechanism and execution uncertainty, which illustrates how the same interaction semantics (the outcome) can induce different execution dynamics under different interaction contexts.

\paragraph{Observation~4: Long-term value amplifies advisory benefit.}
In the corridor, the optimal advice remains unchanged, yet both the advisory advantage over silence and the optimal value increase steadily with attention. 
This shows that advisory effectiveness depends not only on agreement, but also on the long-term value associated with accepted proposals.

\section{Related Work and Modeling Positioning}
\label{sec:related_work}

% \paragraph{Shared Autonomy and Human-in-the-Loop Planning.}
\noindent \textbf{Shared Autonomy and Human-in-the-Loop Planning.} \ 
Shared autonomy studies how control is distributed between human and AI agents, often via switching between teleoperation and full autonomy~\cite{BasichSWWBZ20,DuchettoKIH18,RigterLH20,RosenthalV12}. 
These approaches treat control as an externally selected mode and assume human availability, whereas in~\cite{VanheeJM19}, Advice-MDPs instead adapt the robot's model to human preferences so its behavior aligns with operator intent. 
Recent work jointly selects robot actions and allocates human assistance under uncertainty over operator performance~\cite{CostenGHL24}.

% \subsection{Multi-Agent and Interactive Decision Models}

% \paragraph{Multi-Agent and Interactive Decision Models.}
\noindent \textbf{Multi-Agent and Interactive Decision Models.} \ 
Multi-agent MDPs and Dec-POMDPs~\cite{boutilier1996planning,dolgov2004graphical,oliehoek2016concise,PolHOW22} model coordination through joint actions that directly induce transitions and rewards, with interaction semantics encoded implicitly in the decision dynamics.

Prior work on collaborative decision-making using action suggestions, as in~\cite{asmar2022collaborative}, assumes that agents share a common objective.
Under this assumption, suggestions provide information that updates the receiving agent's belief, rather than being subject to an explicit evaluation mechanism governed by potentially different objectives.
In contrast, our framework explicitly separates interaction evaluation from execution, allowing proposals to be evaluated under another entity's context-dependent objectives, knowledge, or decision model.

Modified-Action MDPs~\cite{langlois2021rl} distinguish actions selected by an agent's policy from those ultimately executed through an action-modification mechanism, primarily to study how reinforcement learning algorithms respond to such modifications.

While sharing this distinction between selected and executed actions, our framework addresses a different modeling question:
how interaction-level proposals are interpreted under context and resolved through evaluation principles that may involve another entity's objectives or perspectives.

The interpretation function makes these principles explicit through semantic outcomes $\sigma$, each admitting application-specific evaluation mechanisms.
This provides a structured representation for specifying and organizing interaction semantics independently of their induced execution dynamics.

The distinction is therefore not one of expressive power, but of modeling focus:
rather than analyzing learning under action modification, our framework makes the principles governing interaction evaluation and resolution explicit within the decision-process specification.

% \subsection{Human-Centered and Cognitive Models}
% \paragraph{Human-Centered and Cognitive Models.}
\noindent \textbf{Human-Centered and Cognitive Models.} \ 
Human decision making is often modeled as context-dependent and resource-bounded, as reflected in theories of bounded rationality~\cite{simon1957behavioral}, dual-process reasoning~\cite{kahneman2011thinking}, and heuristic decision making~\cite{chaiken1980heuristic}. 
These perspectives emphasize that evaluation depends on available cognitive resources and context.

Theory-of-Mind and human-aware planning model agent beliefs and perspectives explicitly~\cite{langley2022theory}.

Our work is complementary: we introduce interaction semantics as a first-class representation of action evaluation and resolution, which can incorporate belief-based or cognitive models when needed~\cite{lourencco2020whose,epley2004perspective}.
% v1

\noindent \textbf{Modeling Positioning.} \ 
Our work is motivated by shared and adjustable autonomy~\cite{Zilberstein15,zieba2010principles,mostafa2019adjustable}, where interaction is typically modeled through extended state variables and control allocation.

Among existing shared-autonomy formulations, the closest is SA-RAL-MDP~\cite{VanheeJM21}, which represents autonomy and support within an MDP and captures behaviors such as authorization, denial, and override through state-dependent transitions.

The difference is representational and structural. 
In SA-RAL-MDP and related approaches, interaction effects are encoded implicitly in transition dynamics, often requiring specification at the level of state-action pairs. 
In contrast, we introduce an explicit semantic layer that maps joint actions and interaction context to semantic outcomes, thereby making interaction semantics explicit before execution.

This induces a factorization in which transition, reward, and discount are defined through evaluation mechanisms associated with semantic outcomes.
As a result, interaction rules can be specified, modified, and potentially learned at the level of semantic outcomes rather than redistributed across transition definitions.

Thus, while overlapping behaviors can be represented in both frameworks, our formulation exposes interaction semantics as a first-class modeling construct.
% , modular, and extensible semantic mechanism for interaction.

\section{Discussion}
\label{sec:dis_and_conclusion}

We introduced a semantics-aware decision framework that explicitly separates the interpretation of interaction-level actions from their induced execution dynamics. 
Rather than embedding evaluation and execution within state-action transition definitions, the framework provides a dedicated semantic layer in which interaction principles are identified and instantiated through application-specific evaluation mechanisms. 
This constitutes a representational distinction while preserving compatibility with standard MDP planning.

A key structural consequence is that distinct joint action-context pairs may be associated with the same semantic outcome, allowing them to invoke a common evaluation mechanism while retaining state- and action-dependent dynamics.
In standard MDP formulations, identifying such commonality requires the modeler to distinguish evaluation from execution and recognize shared evaluation principles within the transition dynamics.
Our framework makes this separation an explicit modeling principle, providing a structured interface for specifying, organizing, and modifying evaluation mechanisms across different interactions.

The shared-autonomy instantiation illustrates how context-dependent evaluation, particularly through attention, influences interaction behavior and optimal decisions.
Although a suitably factored MDP can realize equivalent dynamics, the semantic representation makes the underlying interaction structure explicit rather than leaving its identification to the modeler.
Quantifying the resulting advantages in specification complexity and modification effort remains an important direction for future work.

The framework requires identifying semantic outcomes and specifying their corresponding evaluation mechanisms, which may be challenging in complex domains. 
The present instantiation assumes observable interaction context and an application-specific acceptance model. 
More elaborate evaluation mechanisms could involve reasoning about another agent's beliefs and objectives, exchanging explanations, or engaging in further interaction to influence how a proposal is evaluated.
Formalizing such multi-stage evaluation processes remains an interesting future direction.

Further work also includes learning evaluation mechanisms from data, extending the framework to partially observable interaction contexts, and developing richer models of interaction evaluation and resolution.

\subsection*{Acknowledgments}
This work was supported by the AMI CMA France 2030 NORMANTHIIA project.

% \input{ReproducibilityChecklist.tex}

%Bibliography
\bibliographystyle{unsrt}  
\bibliography{references}

\end{document}